\documentclass[10pt,twocolumn,letterpaper]{article}

\usepackage{cvpr}

\usepackage{amsmath,amsfonts,bm}

\def\eqref#1{equation~\ref{#1}}

\def\1{\bm{1}}

\DeclareMathAlphabet{\mathsfit}{\encodingdefault}{\sfdefault}{m}{sl}
\SetMathAlphabet{\mathsfit}{bold}{\encodingdefault}{\sfdefault}{bx}{n}

\usepackage{multirow}
\usepackage{placeins}
\usepackage{microtype}

\definecolor{cvprblue}{rgb}{0.21,0.49,0.74}
\usepackage[breaklinks,colorlinks,allcolors=cvprblue]{hyperref}

\renewcommand{\topfraction}{0.50}
\renewcommand{\textfraction}{0.15}
\renewcommand{\floatpagefraction}{0.75}

\def\confName{CVPR}
\def\confYear{2027}

\newcommand{\lewm}{\textsc{LeWM}}

\title{AVL-JEPA: Preventing Causal Dynamics Information Collapse in Joint Embedding Predictive Architecture World Models}

\author{Yikang Qiao$^{1}$ \quad Ling Zhang$^{1,*}$ \quad Ziying Song$^{2,*}$ \quad Duan Huang$^{1}$\\%
{\normalsize $^{1}$Central South University \qquad $^{2}$Yanshan University}\\
{\normalsize $^{*}$Corresponding authors}}

\begin{document}
\maketitle

\begin{abstract}
Joint embedding predictive architectures (JEPAs) predict future latent representations without reconstructing observations, enabling world models to focus on high-level semantic dynamics. However, a JEPA can preserve high dimensional visual information while discarding information about the physical consequences of actions. We call this failure mode \emph{causal dynamics information collapse} and propose action-grounded vision-invariance latent (AVL) to prevent this collapse. We first use the executed action as an auxiliary dynamics anchor that encourages the model to preserve dynamics information, and then use a vision-invariance pathway which aligns perturbed and clean latent predictions without discarding dynamics information, forcing the model to fully understand and utilize causal dynamics information. We validate AVL on four robotic control tasks (TwoRoom, PushT, OGBench Cube, and Reacher), showing that it substantially improves success rates under visual perturbations while preserving clean-environment performance. We further evaluate physical consequence alignment, clean--noisy dynamics consistency, and the causal effect of targeted transition subspace erasure. Collectively, these results indicate that dynamic information causally relevant to planning is preserved from collapse under AVL.
\end{abstract}

\begin{figure}[!t]
\centering
\includegraphics[width=0.98\linewidth]{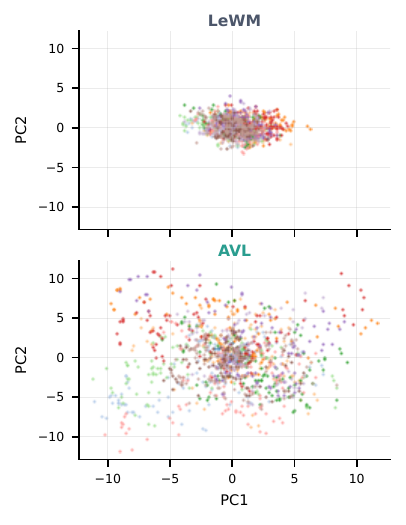}
\caption{Measured PCA projection of predicted $\Delta z$ transition responses for \lewm{} and AVL (ours). Both panels use one shared transform and identical action candidates; different colors denote different candidate action categories.}
\label{fig:pca}
\end{figure}

\section{Introduction}

World models learn to predict how an environment evolves, providing a compact substrate for planning and control \citep{ha2018worldmodels}. Joint-embedding predictive architecture (JEPA) offers one way to achieve this goal. Instead of reconstructing future observations, a JEPA model predicts future latent representations directly \citep{assran2023ijepa}. This design sidesteps pixel-level reconstruction and focuses on learning abstract predictive structure, making it attractive for world modeling.

A central challenge in JEPA training is preventing representation collapse, where all inputs map to a narrow region of latent space. Existing work addresses this at the representation level: variance, covariance, and distributional regularizers penalize degenerate latent statistics \citep{bardes2022vicreg}; EMA-based teacher networks provide a slowly moving prediction target that resists instantaneous collapse \citep{assran2023ijepa,bardes2024vjepa}; JEPA world models such as \lewm{} combine joint-embedding prediction with SIGReg regularization to prevent geometry-degenerate latent embeddings \citep{maes2026lewm,balestriero2025lejepa}. These mechanisms are effective at preventing global collapse and make the learned representations remain high dimensional, non-constant, and predictive on average.

However, preventing global collapse does not guarantee that the latent preserves the information needed to distinguish how different actions causally change the physical state. Two distinct actions that produce different physical state changes may yield nearly identical predicted latent transitions. We refer to this failure as \emph{causal dynamics information collapse}.

We introduce action-grounded vision-invariance latent (AVL) to mitigate this collapse. The action-grounded pathway treats the real action as an auxiliary dynamic anchor, forcing the model to preserve enough dynamic information. The vision-invariance pathway then exposes the model to clean and visually perturbed observations while freezing the action head. The perturbed latent is aligned with the clean latent, but the action loss prevents this alignment from being achieved by simply discarding information, making the model fully use causal dynamics information to predict the future latent.

Our method substantially mitigates causal dynamics information collapse and improves closed-loop planning performance. On TwoRoom, we lift predicted-to-physical consequence alignment (Spearman $\rho$) from $0.35$ to $0.53$ and reduce normalized regret from $0.152$ to $0.061$, while clean--noisy ranking consistency increases from $0.89$ to $0.98$. In closed-loop planning, success rate under Gaussian policy input noise ($\sigma{=}0.10$) improves from $15.3\%$ to $81.5\%$ on PushT and from $65.7\%$ to $88.7\%$ on TwoRoom. Under saturation shift ($+0.35$), it increases from $14.0\%$ to $62.0\%$ on PushT and from $54.8\%$ to $86.7\%$ on TwoRoom. We also improve OGBench Cube (from $73.2\%$ to $78.2\%$ under clean) and remain competitive on Reacher. These results suggest that global non-collapse is only a starting point, a useful world-model latent should preserve causal dynamics-relevant physical structure.

Our main contributions are:
\begin{itemize}
    \item We formulate \emph{causal dynamics information collapse}, a failure mode in which high dimensional visual latents insufficiently preserve action-conditioned physical change, and operationalize it with controlled experiments tied to physical outcomes.
    \item We introduce action-grounded vision-invariance latent (AVL) that targets the collapse: the action-grounded pathway grounds state-conditioned latent transitions with executed actions to preserve dynamic information in latent space, and the vision-invariance pathway aligns perturbed and clean latents while freezing the action head, forcing the model to understand and utilize causal dynamics information to predict future latent.
    \item We validate AVL across TwoRoom, PushT, Reacher, and OGBench Cube, showing that it substantially reduces success-rate degradation under Gaussian, brightness, and saturation perturbations while preserving clean environment performance, and provide mechanism-level experiments showing that the preserved dynamic information is causally meaningful.
\end{itemize}

\section{Related Work}

\paragraph{Latent world models.}
World models learn a compressed simulation of the environment for planning and control \citep{ha2018worldmodels}. Early latent dynamics models reconstruct observations from learned states and train policies in latent imagination \citep{hafner2019latent,hafner2020dreamer,hafner2025dreamerv3,wu2023daydreamer}. Reconstruction-free alternatives predict future latent features directly: DINO-WM plans on top of frozen DINOv2 features \citep{zhou2025dinowm,oquab2024dinov2}, PLDM combines joint-embedding prediction with VICReg-style collapse prevention and inverse dynamics \citep{sobal2025learning}, and \lewm{} introduces SIGReg regularization for stable end to end training from pixels \citep{maes2026lewm,balestriero2025lejepa}.

\paragraph{Joint embedding predictive architecture.}
JEPA predicts target embeddings from context embeddings without pixel reconstruction and has become a prominent framework for self supervised representation learning \citep{lecun2022path,assran2023ijepa}. A central challenge in JEPA training is representation collapse, where all inputs map to a narrow latent region. Early preventions rely on architectural asymmetry: I-JEPA uses an EMA target encoder and masked context prediction so that the context encoder cannot trivially copy the target \citep{assran2023ijepa}. V-JEPA extends this to video with spatiotemporal masking \citep{bardes2024vjepa}, and V-JEPA 2 scales the same recipe to large-scale video understanding and planning \citep{assran2025vjepa2}. Others use explicit statistical regularizers: VICReg penalizes latent variance, invariance, and covariance \citep{bardes2022vicreg}, and LeJEPA provides a provable non-collapse guarantee through spectral regularization \citep{balestriero2025lejepa}. \lewm{} uses SIGReg to maintain non-degenerate latent statistics in world model \citep{maes2026lewm}, Sub-JEPA adds subspace Gaussian regularization \citep{zhao2026subjepa}, and SD-JEPA decomposes the latent into progression and content subspaces \citep{thil2026sdjepa}. Concurrent works such as SMWM and Delta-JEPA expect the inverse-dynamics method alone to prevent collapse \citep{ivashkov2026smwm,zhang2026deltajepa}. However, these methods mostly address collapse at the level of global latent statistics: variance, rank, covariance, or distributional non-degeneracy. Our work targets this gap: we study how to prevent the selective loss of causal dynamics information in JEPA world models, an information collapse directly limits model performance.

\section{Method}
\label{sec:method}

\begin{figure*}[t]
\centering
\includegraphics[width=\linewidth]{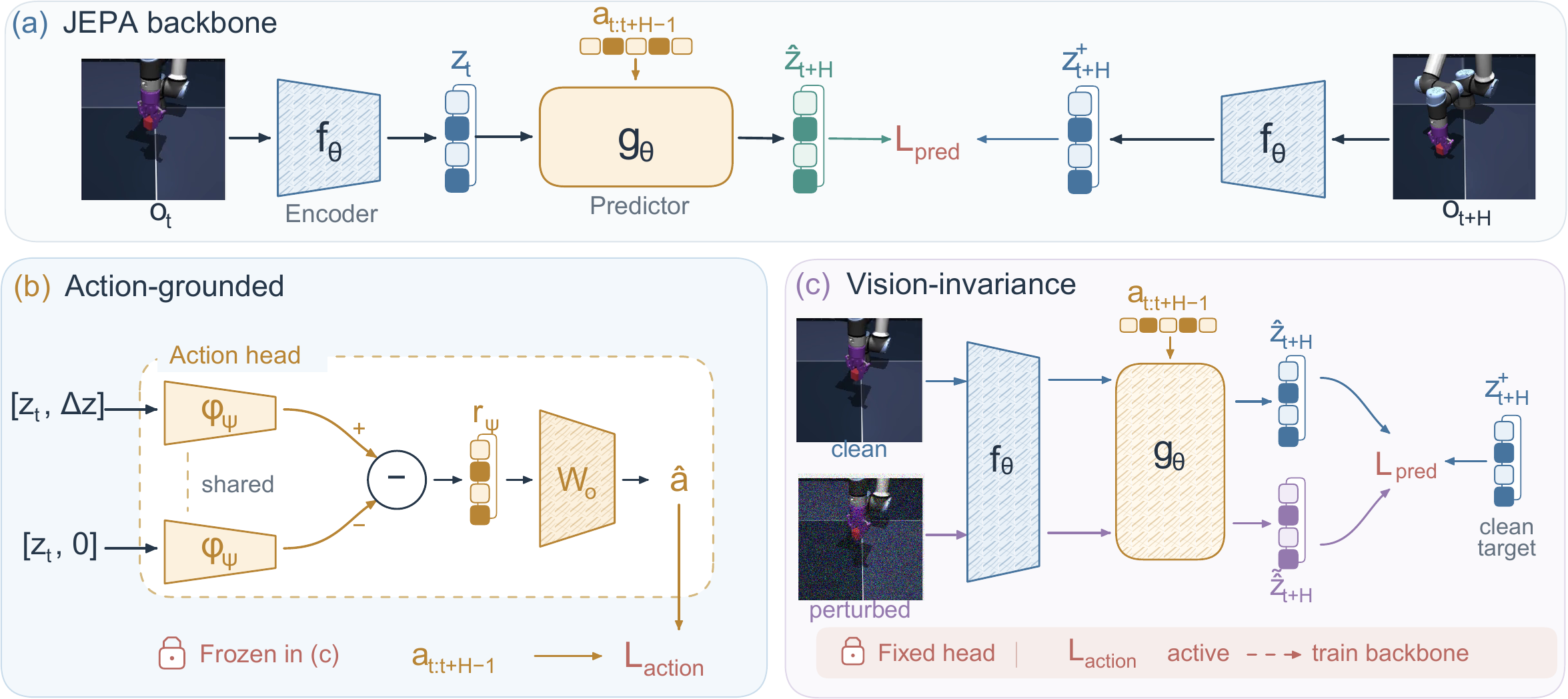}
\caption{Overview of the proposed AVL method. The action-grounded pathway anchors clean latent transitions to executed actions through a transition-attributed action head that subtracts a state-only zero-transition baseline. The vision-invariance pathway aligns noisy and clean latent predictions while freezing the action head. The alignment creates a potential collapse shortcut, but the frozen action loss requires the latent to retain action-relevant dynamics, thereby targeting causal dynamics information collapse.}
\label{fig:method}
\end{figure*}

\subsection{Causal Dynamics Information Collapse}

The JEPA prediction objective rewards information according to its predictability, which need not coincide with its causal relevance for control. Let $X_t$ denote the information available to the predictor, let $u_t$ be a candidate component of the current representation, and let $Y_{t+H}$ denote the vector-valued future latent target. Under squared prediction loss, define the Bayes risks with and without retaining $u_t$ as
\begin{equation}
\mathcal R(X_t)=\mathbb E\!\left[\left\|Y_{t+H}-\mathbb E[Y_{t+H}\mid X_t]\right\|_2^2\right].
\label{eq:bayes_risk_context}
\end{equation}
\begin{equation}
\mathcal R(X_t,u_t)=\mathbb E\!\left[\left\|Y_{t+H}-\mathbb E[Y_{t+H}\mid X_t,u_t]\right\|_2^2\right].
\label{eq:bayes_risk_context_component}
\end{equation}
The predictability gain from retaining $u_t$ is therefore
\begin{equation}
\begin{aligned}
\Delta\mathcal R(u_t)&=\mathcal R(X_t)-\mathcal R(X_t,u_t) \\
&=\mathbb E\!\left[\operatorname{tr}\operatorname{Var}\!\left(\mathbb E[Y_{t+H}\mid X_t,u_t]\mid X_t\right)\right]\geq 0.
\end{aligned}
\label{eq:predictability_gain}
\end{equation}
Thus, temporally persistent visual factors can be favored because they directly reduce future-prediction uncertainty, whereas action-induced physical changes may require nonlinear state--action interactions and provide a weaker or less direct optimization signal. This creates a route by which visual information is retained while action-relevant dynamics are underrepresented.

The JEPA prediction objective also identifies inputs only up to predictive equivalence. For an action set $\mathcal A$, define
\begin{equation}
\begin{split}
 o\sim_{\mathcal A}o'
 \quad\Longleftrightarrow\quad
 &p(z_{t+H}^{+}\mid o,a)=p(z_{t+H}^{+}\mid o',a),\\
 &\forall a\in\mathcal A.
\end{split}
 \label{eq:predictive_equivalence}
\end{equation}
JEPA prediction does not require states within one such equivalence class to remain distinguishable, even if they have different physical consequences under actions outside the observed training support. Consequently, generic non-collapse constraints can maintain non-degenerate latent geometry without guaranteeing causal dynamics representations.

Regularizers such as SIGReg and VICReg act on the marginal distribution of the
encoder outputs: over a batch of representations they penalize degenerate variance, correlated
dimensions and deviation from a target distribution, and they do so without conditioning on the
action that generated each transition. The two quantities that JEPA prediction leaves
underconstrained, however, are conditional ones: how the difference between successive latents
depends on the candidate action, and whether states in the equivalence class of
Equation~\ref{eq:predictive_equivalence} remain distinguishable. An encoder whose latent is a
function of the observation alone, and which carries no information about dynamics,
can still satisfy the marginal penalties whenever the observations themselves are diverse,
because its output statistics look healthy on that batch. A penalty on the marginal therefore
cannot detect the loss of a condition. 

\subsection{AVL Overview}

Our method intervenes at two points in training. The action-grounded pathway establishes a clean dynamics-aware latent space using an action anchor. The vision-invariance pathway uses visual invariance as an alignment constraint: the latent transition predicted from a perturbed observation is aligned with the corresponding clean transition. Because the two views naturally differ, this alignment can be achieved either by preserving and utilizing the relevant causal dynamics or by discarding information until the two latent become similar. The frozen action head and its active loss rule out the latter shortcut, so the model is driven to infer state changes from causal dynamics.

\subsection{Action-grounded: transition-attributed action anchoring}

Let $o_t$ be the current observation, $o_{t+H}$ the observation after $H$ low level action steps, $a_{t:t+H-1}$ the executed action sequence, $z_t=f_\theta(o_{\leq t})$ the current latent produced by the encoder, $z_{t+H}^{+}$ the target future latent produced by the same shared encoder, and $\Delta z=z_{t+H}^{+}-z_t$. Here $H$ is the prediction horizon and the superscript $+$ denotes the target branch. The action-conditioned predictor $g_\theta$ produces $\hat z_{t+H}=g_\theta(z_t,a_{t:t+H-1})$.

The latent may contain task-irrelevant information, so a raw latent difference does not necessarily isolate the dynamics relevant to action prediction. The shared extractor provides feature space in which the transition response can be attributed relative to the same state, while the zero-transition branch removes the state-only component. Specifically, the action head uses a three-layer Linear--LayerNorm--GELU trunk, with input dimension $2d_z$, followed by a bias-free output projection. We construct its state-conditioned transition response as
\begin{equation}
    r_\psi(z_t,\Delta z)=\phi_\psi([z_t,\Delta z])-\phi_\psi([z_t,\mathbf 0]),
    \label{eq:response}
\end{equation}
where $\phi_\psi$ is the shared trainable trunk, $[z_t,\Delta z]$ denotes concatenation, and $\mathbf 0$ is a zero transition of the same dimension as $\Delta z$. So it removes the state-only component and makes $r_\psi$ the transition-attributed response. We then predict the executed action sequence with
\begin{equation}
    \hat a_{t:t+H-1}=W_o r_\psi(z_t,\Delta z),
    \label{eq:action_head}
\end{equation}
where $W_o\in\mathbb{R}^{d_a\times h}$ is the bias-free output projection from the transition-attributed hidden response to the task-specific $H$-step action target. In the implementation, the trunk is $\operatorname{Linear}(2d_z,h)\mathbin{\to}\operatorname{LayerNorm}(h)\mathbin{\to}\operatorname{GELU}$ followed by two $\operatorname{Linear}(h,h)\mathbin{\to}\operatorname{LayerNorm}(h)\mathbin{\to}\operatorname{GELU}$ blocks, with $d_z{=}192$ and $h{=}256$, while $d_a$ is task-specific. Because the identical zero-transition branch is subtracted, $r_\psi(z_t,\mathbf 0)=\mathbf 0$ exactly; this property does not require the hidden trunk itself to be bias-free.

The action-grounded objective is
\begin{equation}
    \mathcal L_{\text{S1}}=\mathcal L_{\text{pred}}+\lambda_1\mathcal L_{\text{SIGReg}}+\lambda_2\mathcal L_{\text{action}},
    \label{eq:action_grounded}
\end{equation}
where $\mathcal L_{\text{pred}}$ is the latent prediction MSE, $\mathcal L_{\text{SIGReg}}$ is the non-collapse regularizer, and
\begin{equation}
    \mathcal L_{\text{action}}=\operatorname{SmoothL1}\!\left(\hat a_{t:t+H-1},a_{t:t+H-1}\right).
\end{equation}

$H$ corresponds to low level action steps after action blocking. The output dimension $d_a$ is task-specific and is determined by the environment action dimension multiplied by the action-block length. The auxiliary branch is trainable together with the encoder, projector, action encoder, and predictor during the action-grounded pathway. It is saved separately and excluded from the exported planning checkpoint.

The action-grounded pathway is evaluated by asking whether its predicted candidate-transition geometry preserves physical consequences. For state $s$, let $\Delta\hat z_s^{(k)}$ be the predicted transition response for candidate $k$, $\Delta x_s^{(k)}$ its task-relevant physical response, $p_s^{(k)}=\lVert\Delta\hat z_s^{(k)}\rVert_2$ the model score, and $q_s^{(k)}=\lVert\Delta x_s^{(k)}\rVert_2$ the physical score. Physical consequence alignment compares all unordered candidate pairs:
\begin{equation}
\begin{split}
 &\rho_{\mathrm{phys}}(s)=\rho_{\mathrm{S}}\Big(
 \{\lVert\Delta\hat z_s^{(i)}-\Delta\hat z_s^{(j)}\rVert_2\}_{i<j},\\
 &\hspace{1.2em}\{\lVert\Delta x_s^{(i)}-\Delta x_s^{(j)}\rVert_2\}_{i<j}\Big),
\end{split}
 \label{eq:physical_spearman}
\end{equation}
where $\rho_{\mathrm{S}}$ is Spearman's rank correlation. With $k_s^*=\arg\max_k q_s^{(k)}$ and $\hat k_s=\arg\max_k p_s^{(k)}$, normalized regret and Top-$K$ oracle inclusion are
\begin{equation}
\begin{split}
 R_{\mathrm{norm}}(s)&=\frac{q_s^{(k_s^*)}-q_s^{(\hat k_s)}}{q_s^{(k_s^*)}+\epsilon},\\
 A_K(s)&=\mathbf 1\!\left[k_s^*\in\operatorname{TopK}\{p_s^{(k)}\}_{k=1}^{M}\right].
\end{split}
 \label{eq:selection_metrics}
\end{equation}
We report state means for $\rho_{\mathrm{phys}}$, $R_{\mathrm{norm}}$, and $A_K$, with $K\in\{1,3\}$. To test whether consequence-ranking information is concentrated in the predicted response, let $U_k$ contain the leading $k$ global principal directions of centered candidate-transition responses. Targeted erasure produces
\begin{equation}
\begin{split}
 \Delta\hat z_{s,-k}^{(i)}&=\Delta\hat z_s^{(i)}-U_kU_k^\top\Delta\hat z_s^{(i)},\\
 \Delta\rho(k)&=\bar\rho(0)-\bar\rho(k),\quad
 \Delta R(k)=\bar R(k)-\bar R(0),
\end{split}
 \label{eq:erasure_metrics}
\end{equation}
where bars denote means over matched states; positive $\Delta\rho$ and $\Delta R$ indicate worse physical ordering after erasure. Finally, for seed $j$ with $E_j$ closed-loop episodes, $S_j=E_j^{-1}\sum_{e=1}^{E_j}\mathbf 1[\text{success}_{j,e}]$, and we report
\begin{equation}
 \bar S=\frac{1}{J}\sum_{j=1}^{J}S_j,\qquad
 \sigma_S=\sqrt{\frac{1}{J}\sum_{j=1}^{J}(S_j-\bar S)^2},
 \label{eq:closed_loop_stats}
\end{equation}
with $J=10$ and population standard deviation ($\mathrm{ddof}=0$).

\begin{figure*}[t]
\centering
\includegraphics[width=0.66\linewidth]{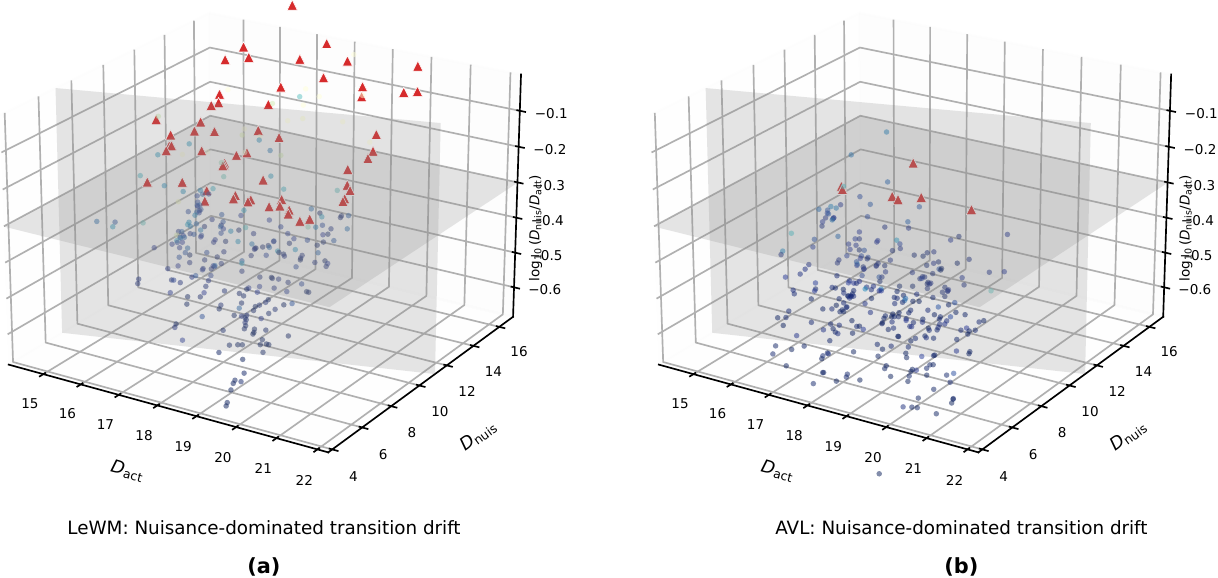}
\caption{Clean--noisy candidate-transition geometry on TwoRoom for \lewm{} and AVL (ours). Red triangles indicate states satisfying $D_{\mathrm{nuis}}/D_{\mathrm{act}}>0.5$ together with a clean--noisy Top-1 candidate flip; blue dots denote all remaining states. The transparent planes show the same ratio threshold in the $(D_{\mathrm{act}},D_{\mathrm{nuis}})$ and $\log_{10}(D_{\mathrm{nuis}}/D_{\mathrm{act}})$ views.}
\label{fig:cdic_diagnostics}
\end{figure*}

\subsection{Vision-invariance: clean--perturbed latent alignment with frozen dynamics anchor}
The vision-invariance pathway applies visual perturbation to a single branch during training: each current input is randomly drawn, with roughly equal probability ($1/3$ each), from $\{$clean, Gaussian noise with $\sigma=0.01$, Gaussian noise with $\sigma=0.03\}$. The perturbation is applied only to the current observation $o_t$; the future target observation $o_{t+H}$ remains clean. Let $z_t=f_\theta(o_{\leq t})$ and $\tilde z_t=f_\theta(\tilde o_{\leq t})$ denote the clean and perturbed context latents for the same underlying state. Their predicted future latents are $\hat z_{t+H}=g_\theta(z_t,a_{t:t+H-1})$ and $\hat{\tilde z}_{t+H}=g_\theta(\tilde z_t,a_{t:t+H-1})$. Both are trained toward the same clean target $z_{t+H}^{+}$, which induces the model to better utilize causal dynamics information.

The consistency constraint only requires pairs that share the same underlying semantics but
differ in pixels, and Gaussian noise on the observation is the mildest perturbation with that
property: it alters the input locally without changing what the scene contains or how it
evolves. The levels $\sigma=0.01$ and $\sigma=0.03$ change the pixels appreciably while leaving
the transition essentially unchanged; larger magnitudes push the input outside the range of
observations seen in training, and the consistency term degrades into fitting the perturbation
itself.
The vision-invariance pathway uses the same loss terms and hyperparameters as the action-grounded pathway:
\begin{equation}
    \mathcal L_{\text{S2}}=\mathcal L_{\text{pred}}+\lambda_1\mathcal L_{\text{SIGReg}}+\lambda_2\mathcal L_{\text{action}},
    \label{eq:vision_invariance}
\end{equation}
The action-grounded action-head parameters are frozen in the vision-invariance pathway, but the action loss remains active and back-propagates through the model.
The vision-invariance pathway is evaluated on matched clean and perturbed predictions. For state $s$, candidate $k$, and perturbation repeat $r$, let $\Delta\hat z_{s,\mathrm c}^{(k)}$ and $\Delta\hat z_{s,r,\eta}^{(k)}$ denote the clean and noisy predicted transitions. The clean action spread and fixed-action nuisance drift are
\begin{equation}
 D_{\mathrm{act}}(s)=\frac{2}{M(M-1)}\sum_{i<j}\left\lVert\Delta\hat z_{s,\mathrm c}^{(i)}-\Delta\hat z_{s,\mathrm c}^{(j)}\right\rVert_2,
 \label{eq:dact_metric}
\end{equation}
\begin{equation}
\begin{split}
 D_{\mathrm{nuis}}(s,r)&=\frac{1}{M}\sum_{k=1}^{M}\left\lVert\Delta\hat z_{s,r,\eta}^{(k)}-\Delta\hat z_{s,\mathrm c}^{(k)}\right\rVert_2,\\
 Q_{\mathrm{A/N}}(s,r)&=\frac{D_{\mathrm{act}}(s)}{D_{\mathrm{nuis}}(s,r)+\epsilon}.
\end{split}
 \label{eq:dnuis_metric}
\end{equation}
Let $d_{s,\mathrm c}^{(ij)}=\lVert\Delta\hat z_{s,\mathrm c}^{(i)}-\Delta\hat z_{s,\mathrm c}^{(j)}\rVert_2$ and $d_{s,r,\eta}^{(ij)}=\lVert\Delta\hat z_{s,r,\eta}^{(i)}-\Delta\hat z_{s,r,\eta}^{(j)}\rVert_2$ for $i<j$. Clean--noisy pairwise-geometry consistency and Top-1 stability are
\begin{equation}
\begin{split}
 C_{\rho}(s,r)&=\rho_{\mathrm S}\!\left(\{d_{s,\mathrm c}^{(ij)}\}_{i<j},\{d_{s,r,\eta}^{(ij)}\}_{i<j}\right),\\
 C_{\mathrm{Top1}}(s,r)&=\mathbf 1\!\left[\arg\max_k p_{s,\mathrm c}^{(k)}=\arg\max_k p_{s,r,\eta}^{(k)}\right],
\end{split}
 \label{eq:clean_noisy_metrics}
\end{equation}

where $p_{s,\mathrm c}^{(k)}=\lVert\Delta\hat z_{s,\mathrm c}^{(k)}\rVert_2$ and $p_{s,r,\eta}^{(k)}=\lVert\Delta\hat z_{s,r,\eta}^{(k)}\rVert_2$. These quantities separate retained action-conditioned variation from perturbation-induced drift and directly test whether the frozen dynamics anchor preserves candidate geometry under visual corruption.

\begin{table*}[t]
\centering
\caption{Clean--noisy dynamics consistency on TwoRoom and PushT (100 matched states; three noisy repeats; Gaussian $\sigma=0.05$). Higher clean--noisy Spearman, Top-1 agreement, and action-to-nuisance ratio are better; lower nuisance drift and noisy regret are better. Best values within each task and metric are bolded.}
\label{tab:stage2}
\renewcommand{\arraystretch}{1.08}
\begin{tabular*}{\linewidth}{@{\extracolsep{\fill}}llccccc@{}}
\toprule
Task & Model & \shortstack{Clean--noisy\\Spearman $\uparrow$} & \shortstack{Top-1\\agreement (\%) $\uparrow$} & \shortstack{Action-to\\nuisance ratio $\uparrow$} & \shortstack{Nuisance\\drift $\downarrow$} & \shortstack{Noisy\\regret $\downarrow$} \\
\midrule
\multirow{3}{*}{TwoRoom} & \lewm{} & 0.89 & 58 & 2.2 & 8.9 & 0.162 \\
& Action-grounded & 0.90 & 60 & 2.0 & 10.1 & 0.088 \\
& AVL (ours) & \textbf{0.98} & \textbf{76} & \textbf{2.9} & \textbf{6.8} & \textbf{0.069} \\
\midrule
\multirow{3}{*}{PushT} & \lewm{} & 0.809 & 45.3 & 2.35 & 6.15 & 0.387 \\
& Action-grounded & 0.970 & 76.0 & 7.21 & 3.60 & 0.313 \\
& AVL (ours) & \textbf{0.998} & \textbf{93.0} & \textbf{19.23} & \textbf{1.07} & \textbf{0.272} \\
\bottomrule
\end{tabular*}
\end{table*}

\section{Experiments}
\label{sec:experiments}

\subsection{Tasks and Metrics}

\begin{figure*}[t]
\centering
\includegraphics[width=\linewidth]{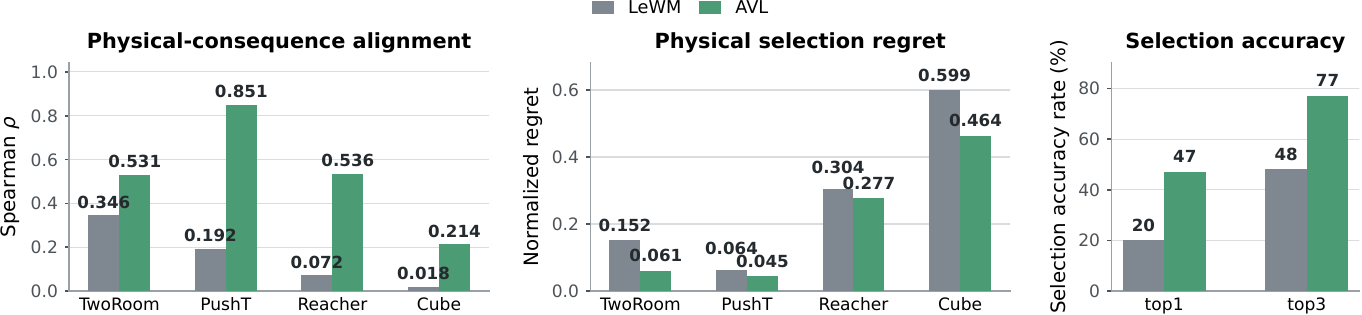}
\caption{Counterfactual physical consequence alignment, selection quality, and TwoRoom selection accuracy. The first panel reports predicted-to-physical Spearman correlation, the second reports normalized regret, and the third reports Top-1 and Top-3 selection accuracy. Higher Spearman and selection accuracy are better, whereas lower regret is better.}
\label{fig:counterfactual_alignment}
\end{figure*}

\begin{table*}[t]
\centering
\renewcommand{\arraystretch}{1.05}
\caption{Clean closed-loop results.}
\label{tab:clean_results}
\begin{tabular*}{\linewidth}{@{\extracolsep{\fill}}lcccc@{}}
\specialrule{0.5pt}{5pt}{0.5ex}
Model & TwoRoom & PushT & Reacher & Cube \\
\specialrule{0.5pt}{0.5ex}{0.5ex}
PLDM & 91.7 $\pm$ 1.3 & 78.1 $\pm$ 1.7 & 78.3 $\pm$ 2.4 & 65.2 $\pm$ 1.5 \\
SD-JEPA & 86.3 $\pm$ 6.2 & 84.6 $\pm$ 4.7 & 85.0 $\pm$ 5.2 & 69.6 $\pm$ 5.3 \\
\lewm{} & 88.4 $\pm$ 2.3 & \underline{96.0 $\pm$ 3.5} & \textbf{86.4 $\pm$ 4.4} & \underline{73.2 $\pm$ 6.8} \\
\specialrule{0.5pt}{0.5ex}{0.5ex}
Action-grounded & \underline{92.8 $\pm$ 2.4} & 95.5 $\pm$ 1.9 & 82.4 $\pm$ 3.9 & 69.1 $\pm$ 5.6 \\
Vision-invariance & 85.2 $\pm$ 3.2 & 91.6 $\pm$ 3.7 & 84.0 $\pm$ 2.8 & 72.4 $\pm$ 7.8 \\
AVL (ours) & \textbf{93.2 $\pm$ 1.6} & \textbf{96.0 $\pm$ 1.3} & \underline{85.2 $\pm$ 4.4} & \textbf{78.2 $\pm$ 5.3} \\
\specialrule{0.5pt}{0.5ex}{0.5ex}
\end{tabular*}
\end{table*}

\begin{figure}[!t]
\centering
\includegraphics[width=\linewidth]{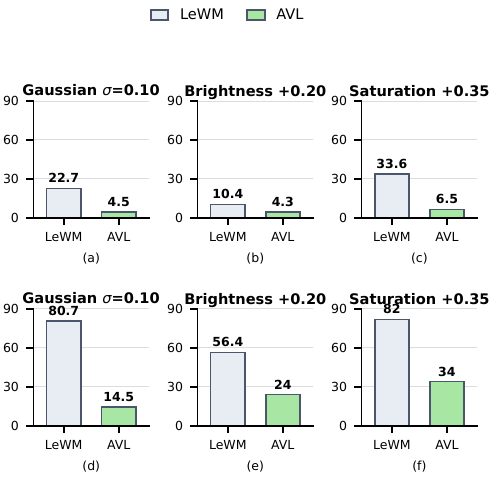}
\caption{Success-rate degradation under Gaussian policy input noise. Panels (a)--(c) report TwoRoom and panels (d)--(f) report PushT.}
\label{fig:visual_drop}
\end{figure}
\label{sec:implementation_details}
\paragraph{Training configurations.}
We train on an NVIDIA RTX 5090 GPU with batch size 128 for all models. The LeWM baseline is trained for 10 epochs with AdamW, learning rate $5\times10^{-5}$, weight decay $10^{-3}$, bf16 mixed precision, and eight data loader workers. The action-grounded pathway is trained from scratch for 10 epochs with the same optimizer settings. The vision-invariance pathway is initialized from the action-grounded weights and trained for 2 epochs with AdamW, learning rate $10^{-5}$, the same weight decay and precision. Both pathways use $\lambda_1=0.09$ for SIGReg, matching the original LeWM configuration, and $\lambda_2=5$ for the action loss.  On TwoRoom, the action-grounded sweep gives success rates of $85.2\%$, $87.2\%$, $92.8\%$, and $92.6\%$ for $\lambda_2\in\{0.05,0.5,5,10\}$, respectively. This sweep supports the choice $\lambda_2=5$: increasing the action weight from $0.05$ to $5$ improves success by 7.6 percentage points, while $\lambda_2=10$ does not improve further. In the vision-invariance pathway the action head is frozen while its loss remains active. The Vision-invariance baseline used as an ablation reference is initialized from the LeWM weights and trained for 2 epochs with the same clean-target perturbation objective but without the action loss. The measured training time of both pathways is 41.7 minutes per epoch on PushT, 15.3 minutes per epoch on TwoRoom, 53.4 minutes per epoch on Reacher, and 61.9 minutes per epoch on Cube. Detailed training conditions are provided in Appendix~\ref{app:main_training_conditions}.
\paragraph{Planning configuration.}
Closed-loop planning uses the LeWorldModel sampling-based planner with the default cross-task interface. The planner uses the cross-entropy method (CEM) with 300 candidate action sequences per iteration, 30 optimization iterations, and the top 30 candidates as elites. The prediction horizon, receding horizon, and action-block size are all 5, with horizons measured in action blocks; thus the planner predicts five action blocks and executes those 25 low level steps before replanning. Goals are offset by 25 environment steps. Each seed is evaluated over 50 episodes, with a closed-loop execution budget of 50 environment steps per episode. The planning configuration is used for all model comparisons unless stated otherwise. The closed-loop evaluation time is 1.8 minutes per seed on PushT, 1.3 minutes per seed on TwoRoom, 1.9 minutes per seed on Reacher, and 2.3 minutes per seed on Cube.
\subsection{Clean--noisy dynamics consistency}

\paragraph{Tasks.}
We organize the study into four complementary experiments. \textbf{Counterfactual physical consequence alignment} (Figure~\ref{fig:counterfactual_alignment}) asks whether action anchoring improves alignment between latent transitions and physical consequences. \textbf{Clean--noisy dynamics consistency} (Table~\ref{tab:stage2}) tests whether clean--perturbed alignment with the frozen dynamics anchor preserves causal transition structure rather than collapsing it on both TwoRoom and PushT. \textbf{Candidate-transition geometry and targeted erasure} (Figure~\ref{fig:targeted_erasure}) examines the geometry of predicted candidate transitions and tests whether the directions used for consequence ranking are causally important by selectively removing them. Finally, \textbf{closed-loop planning} (Table~\ref{tab:clean_results}) evaluates MPC performance to determine whether our model retains and utilizes causal dynamics information. We evaluate mechanism-level consistency on TwoRoom and PushT, and closed-loop planning on TwoRoom, PushT, Reacher, and OGBench Cube, following standard latent-world-model benchmarks and continuous-control suites \citep{park2025ogbench,tassa2018dmcontrol,tunyasuvunakool2020dmcontrol}.

\paragraph{Experiment evaluation protocol.}
For closed-loop planning, each model is evaluated independently over 10 random seeds, with 50 episodes per seed. The action-grounded auxiliary action head uses a 256-dimensional hidden response and a task-specific output dimension $d_a$ spanning $H=5$ low level action steps. The auxiliary branch is saved separately and filtered from the weights used at evaluation, so planning uses the original \lewm{} architecture and interface with no additional modules. The remaining experiment evaluation protocols and metrics are detailed in Appendix~\ref{app:mechanism_protocols}.

\subsection{Implementation Details}

We analyze the collapse using the clean--perturbed quantities defined in Equations~\ref{eq:dact_metric}--\ref{eq:clean_noisy_metrics}. All quantities are first averaged across the three perturbation repeats for each state and then across states. Figure~\ref{fig:cdic_diagnostics} visualizes the same-state nuisance/action geometry; red markers identify $D_{\mathrm{nuis}}/D_{\mathrm{act}}>0.5$ together with a Top-1 flip.

On TwoRoom, the vision-invariance pathway supplies the clean--perturbed alignment constraint while the frozen action loss prevents the model from discarding action-relevant variation. Relative to Action-grounded, clean--noisy Spearman increases by 0.0808 (95\% CI $[0.0646,0.0987]$), same-top-1 frequency increases by 16.3 percentage points (95\% CI $[9.7,23.0]$), nuisance drift decreases by 3.318 (95\% CI $[-3.579,-3.074]$), and action-to-nuisance ratio increases by 0.899 (95\% CI $[0.840,0.960]$). The independently repeated PushT protocol shows the same stability pattern: AVL has the highest clean--noisy Spearman and Top-1 agreement, the lowest nuisance drift, and the highest action-to-nuisance ratio. The simultaneous reduction in nuisance drift and retention of action separation is consistent with mitigating causal dynamics information collapse rather than uniformly suppressing latent variation.

\subsection{Counterfactual physical consequence alignment}

Figure~\ref{fig:counterfactual_alignment} summarizes counterfactual physical consequence alignment across all four tasks. We report predicted-to-physical Spearman correlation and normalized regret, which directly measure whether the model orders candidate transitions according to their physical consequences and whether its selected candidate is close to the physical oracle. AVL improves both measures on TwoRoom, PushT, Reacher, and Cube in the aggregate view; Appendix~\ref{app:mechanism_analysis} decomposes these gains by training stage.

\begin{figure}[!t]
\centering
\includegraphics[width=\linewidth]{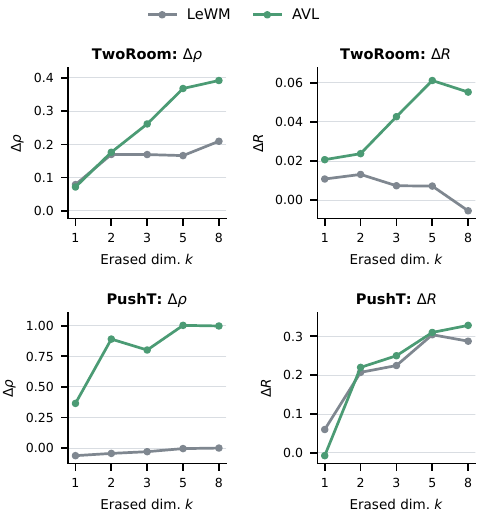}
\caption{Targeted erasure sweep on TwoRoom and PushT. Each panel reports the signed difference between the erased and corresponding non-erased baseline values across $k\in\{1,2,3,5,8\}$.}
\label{fig:targeted_erasure}
\end{figure}

\begin{figure}[t]
\centering
\includegraphics[width=\linewidth]{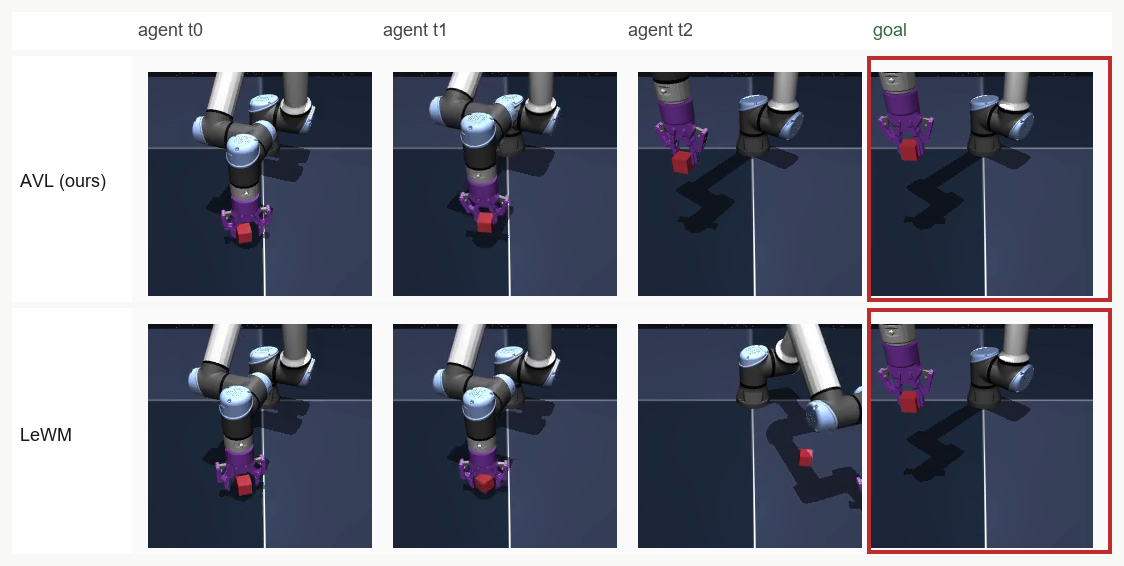}
\caption{Closed-loop OGBench Cube rollouts. The top row shows a successful AVL (ours) rollout, whereas the bottom row shows a failed \lewm{} rollout under the same task. Each row contains three temporally ordered agent observations and the corresponding goal observation.}
\label{fig:task_visualization}
\end{figure}

\subsection{Candidate-transition geometry and targeted erasure}

We apply the same $k$-dimensional leading-subspace erasure procedure to both models following Equation~\ref{eq:erasure_metrics} and compare the resulting changes in physical consequence ranking; a larger performance degradation for AVL indicates that its leading transition-response directions concentrate more information relevant to causal consequence. Figure~\ref{fig:targeted_erasure} reports the complete $k\in\{1,2,3,5,8\}$ sweep on TwoRoom and PushT.

In each panel the horizontal axis is the number of erased leading transition-response
dimensions $k$ and the vertical axis is the signed change of the diagnostic relative to the
non-erased baseline, so a curve moving away from zero means that the erased directions carried
information the diagnostic depends on; the two columns differ only in the diagnostic and the
two rows differ only in the task. Appendix~\ref{app:mechanism_analysis} discusses the sign pattern and the random-subspace control.

\subsection{Main closed-loop results}
Table~\ref{tab:clean_results} reports clean closed-loop results; the corresponding visual-perturbation results are reported in Appendix~\ref{app:visual_conditions}. Each table cell reports the mean success rate and population standard deviation defined in Equation~\ref{eq:closed_loop_stats} across 10 seeds, in percentage points. In Table~\ref{tab:clean_results}, the best result in each task column is bold and the second-best is underlined.

The full AVL (ours) pathway attains the highest success rate on TwoRoom and PushT across all four conditions and on Cube under most brightness shifts. Figure~\ref{fig:visual_drop} reports degradation relative to each model's own clean performance.

Each bar is the drop in success rate of one model under one perturbation, in percentage
points, so a taller bar means a larger loss of competence.

Another observation concerns generalization. The vision-invariance pathway is trained only with Gaussian noise on the policy input observation, yet the model also improves performance on brightness and saturation shifts, indicating the transfer is consistent with learning and exploiting causal dynamics information rather than merely fitting a specific augmentation. Across both tasks, saturation is the hardest of the three conditions for every model in Figure~\ref{fig:visual_drop}, and also the condition under which the degradation of AVL is largest in absolute terms.

\section{Conclusion}

We studied how to prevent causal dynamics information collapse in JEPA world models. Our AVL method uses action anchoring and constrained clean--perturbed alignment to preserve and exploit action conditioned physical structure. Counterfactual consequence ranking, clean--noisy consistency, targeted subspace erasure, and closed-loop transfer across unseen perturbations provide complementary evidence that the method effectively prevents this collapse.

\paragraph{Limitations and future work.}
AVL builds on the SIGReg regularizer of \lewm{} rather than replacing it, and we evaluate it only in simulation. We plan to remove the dependence on regularizers and to conduct real robot experiments in future work.

\clearpage

\subsection*{LLM Usage Statement}
Large language models were used for language editing only. Concretely, we used GPT-6-Astra
and DeepSeek-V4.1-Flash on the abstract, the introduction and the related-work section to
improve grammar, clarity and concision. Every AI-assisted passage was reviewed and edited by
the authors, who take full responsibility for the content of the paper.

{
    \small
\setlength{\bibsep}{0pt plus 0.2ex}
\bibliographystyle{ieeenat_fullname}
    \bibliography{references}
}

\clearpage
\setcounter{page}{1}
\onecolumn

\appendix
\renewcommand{\topfraction}{0.90}
\renewcommand{\textfraction}{0.05}
\renewcommand{\floatpagefraction}{0.75}
\renewcommand{\dbltopfraction}{0.90}
\setcounter{table}{0}
\renewcommand{\thetable}{A\arabic{table}}
\setcounter{figure}{0}
\renewcommand{\thefigure}{A\arabic{figure}}

\section{Notation}
\label{app:notation}

\begin{table}[!h]
\centering
\caption{Summary of the notation used throughout the paper.}
\label{tab:notation}
\renewcommand{\arraystretch}{0.92}
\begin{tabular}{@{}p{0.34\linewidth}p{0.62\linewidth}@{}}
\toprule
Notation & Description \\
\midrule
\multicolumn{2}{@{}l}{\emph{Observations and latents}} \\
$o_t$ & Observation at time step $t$ \\
$\tilde o_t$ & Visually perturbed observation at time step $t$ \\
$o_{t+H}$ & Observation after $H$ low level action steps \\
$z_t=f_\theta(o_{\leq t})$ & Current latent produced by the encoder \\
$\tilde z_t$ & Latent of the perturbed observation \\
$z_{t+H}^{+}$ & Target future latent from the shared encoder \\
$\Delta z=z_{t+H}^{+}-z_t$ & Latent transition \\
$X_t$ & Information available to the JEPA predictor at time $t$ \\
$u_t$ & Candidate component of the current representation \\
$Y_{t+H}$ & Vector-valued future latent target \\
$\mathcal A$ & Action set used to define predictive equivalence \\
$o\sim_{\mathcal A}o'$ & Predictive equivalence under all actions in $\mathcal A$ \\
$\Delta\mathcal R(u_t)$ & Reduction in squared-prediction Bayes risk from retaining $u_t$ \\
$\hat z_{t+H}$ & Predicted future latent \\
$\hat{\tilde z}_{t+H}$ & Predicted future latent from the perturbed input \\
\midrule
\multicolumn{2}{@{}l}{\emph{Actions and predictor}} \\
$a_{t:t+H-1}$ & Executed action sequence over the horizon \\
$\hat a_{t:t+H-1}$ & Predicted action sequence \\
$g_\theta$ & Action-conditioned predictor \\
$\phi_\psi$ & Shared transition-attribution trunk \\
$r_\psi(z_t,\Delta z)$ & Transition-attributed response \\
$W_o$ & Bias-free output projection \\
\midrule
\multicolumn{2}{@{}l}{\emph{Dimensions and hyperparameters}} \\
$H$ & Prediction horizon (action-block length) \\
$d_z$ & Latent embedding dimension \\
$h$ & Hidden trunk dimension \\
$d_a$ & Task-specific action output dimension \\
\midrule
\multicolumn{2}{@{}l}{\emph{Training objectives}} \\
$\mathcal L_{\text{pred}}$ & Latent prediction MSE \\
$\mathcal L_{\text{SIGReg}}$ & SIGReg non-collapse regularizer \\
$\mathcal L_{\text{action}}$ & Action prediction loss \\
$\mathcal L_{\text{S1}}$, $\mathcal L_{\text{S2}}$ & Action-grounded and vision-invariance objectives \\
$\lambda_1$, $\lambda_2$ & SIGReg and action loss weights \\
\midrule
\multicolumn{2}{@{}l}{\emph{Evaluation metrics}} \\
$\rho_{\mathrm{phys}}$ & Predicted-to-physical Spearman correlation \\
$R_{\mathrm{norm}}$ & Normalized regret \\
$A_K$ & Top-$K$ oracle inclusion \\
$D_{\mathrm{act}}$ & Clean action spread \\
$D_{\mathrm{nuis}}$ & Fixed-action nuisance drift \\
$Q_{\mathrm{A/N}}$ & Action-to-nuisance ratio \\
$C_\rho$ & Clean--noisy pairwise-geometry consistency \\
$C_{\mathrm{Top1}}$ & Clean--noisy Top-1 stability \\
$U_k$ & Leading $k$ principal directions of transition responses \\
$\bar S$ & Mean success rate across seeds \\
$\sigma_S$ & Population standard deviation of the success rate \\
$J$ & Number of evaluation seeds \\
$E_j$ & Number of closed-loop episodes for seed $j$ \\
\bottomrule
\end{tabular}
\end{table}

\section{Additional Experimental Details}

\subsection{Mechanism-level protocols}
\label{app:mechanism_protocols}
For the counterfactual experiment, we sample 100 matched simulator states and evaluate the same 12 candidate action sequences at every state, with each candidate spanning 5 action blocks. The simulator executes the shared candidate bank to obtain physical consequences. Spearman correlation and normalized regret measure ordering agreement and the physical cost of model selection; top-$k$ inclusion is reported as a secondary selection metric. For clean--noisy consistency, Gaussian noise with $\sigma=0.05$ is applied only to the current policy observation, while the physical rollout, goal, and success criterion remain clean; three noisy repeats are evaluated per state. Clean--noisy ranking consistency measures whether candidate ordering is preserved; nuisance drift measures perturbation-induced transition change at fixed action, and the action-to-nuisance ratio compares action-induced separation with that drift. Candidate-transition erasure uses the same predicted candidate responses and compares targeted removal of leading principal directions with random orthogonal removal.

\subsection{Counterfactual protocol}
The archived mechanism experiment uses 100 matched simulator states and 12 shared candidate action sequences per state. Each candidate spans five action blocks, and each sequence covers 25 physical low level steps. Predicted latent responses, target latent responses, and two dimensional physical responses are saved for every state--candidate pair. Simulator restoration was audited over 1,200 renders per model with zero non-exact restorations and zero pixel difference. States are sampled from the held-out TwoRoom trajectory pool with a fixed seed; the shared action bank is reused for both models and consists of 12 sequences of five action blocks. Images use the model's standard ImageNet normalization after any evaluation-time uint8 corruption. The physical oracle is the candidate with the largest physical response norm; regret is the oracle-minus-chosen response difference. All erasure branches rescore the same cached candidate responses.

\subsection{PCA projection analysis}
\label{app:pca_analysis}
Figure~\ref{fig:pca} is computed from archived TwoRoom counterfactual responses rather than from a conceptual drawing. For each of 100 matched simulator states, the same bank of 12 candidate action sequences is evaluated by \lewm{}, Action-grounded, and AVL. Each candidate contains five action blocks, corresponding to 25 low level environment steps. The plotted quantity is the predicted transition response $\Delta\hat z=\hat z_{t+H}-z_t\in\mathbb{R}^{192}$, yielding an array of size $100\times12\times192$ for each model. The action bank contains zero, cardinal, diagonal, and three smooth random candidate categories. Different colors in Figure~\ref{fig:pca} denote different candidate action categories; the individual category names are omitted from the main figure to keep the comparison visually compact.

To make the two displayed panels directly comparable, we first concatenate all predicted transition responses from \lewm{}, Action-grounded, and AVL, standardize every latent dimension using one shared \texttt{StandardScaler}, and fit one two-component PCA transform to the resulting pooled matrix. We then project the \lewm{} and AVL responses through this same fitted scaler and PCA transform and display those two panels with identical coordinate limits. Action-grounded responses participate in fitting the shared transform to preserve the original three-model analysis protocol, but its panel is omitted from Figure~\ref{fig:pca} for clarity. The first two components explain 5.80\% and 4.94\% of the standardized variance, respectively. Because the projection retains only two of 192 dimensions, it is used as a qualitative visualization of transition-response organization, not as standalone quantitative evidence of physical consequence alignment or planning performance; those claims are evaluated using the counterfactual ranking, regret, erasure, and closed-loop results.

\subsection{Perturbation protocol}
\begin{figure*}[!t]
    \centering
    \includegraphics[width=0.68\linewidth]{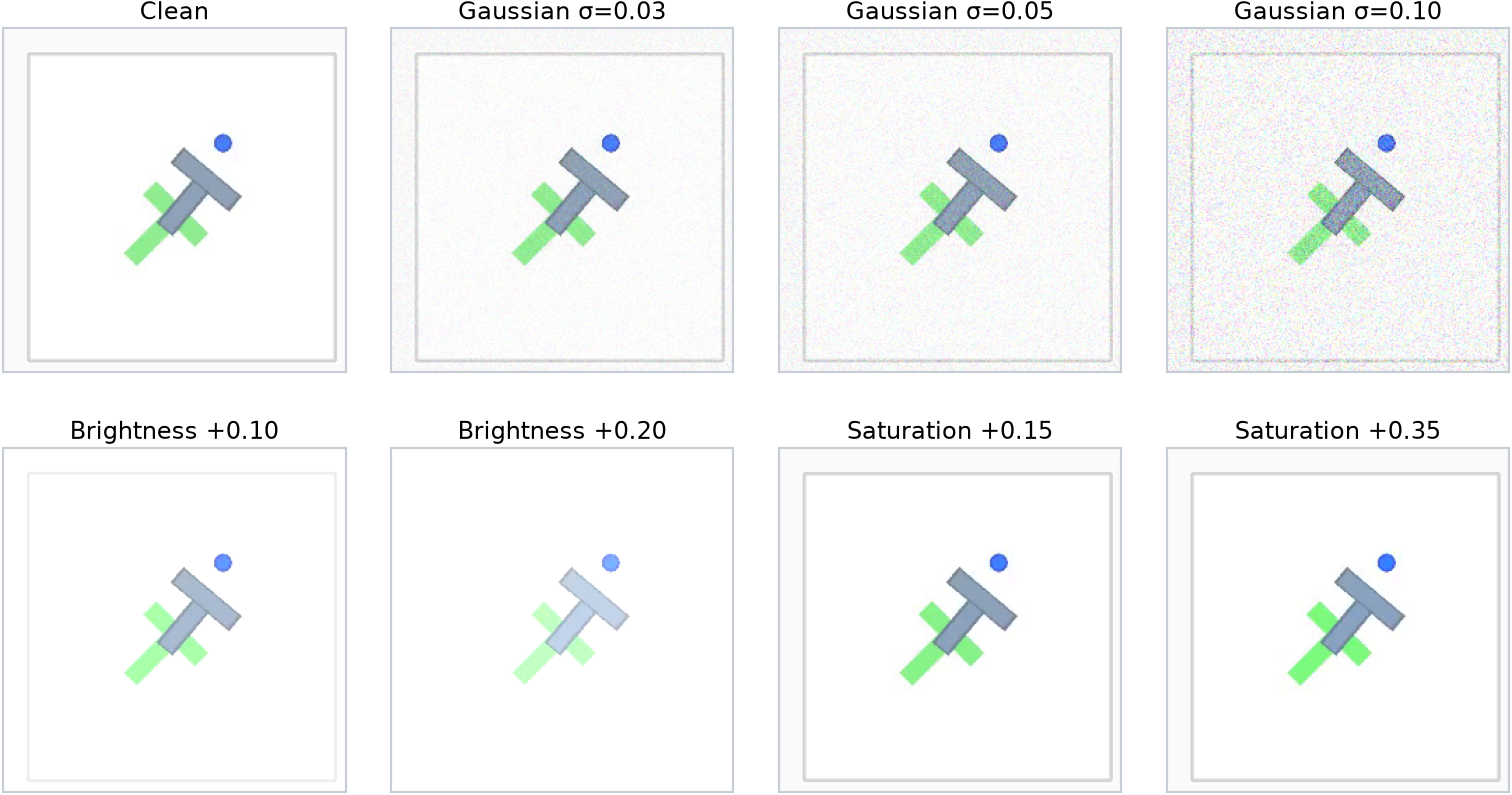}
    \caption{Input-level visual perturbation examples on an archived real PushT RGB frame. The clean image is shown alongside the Gaussian, brightness, and saturation settings used in the evaluation protocol. These examples illustrate the appearance transformation only; they are not episode-level performance measurements.}
    \label{fig:pusht_visual_perturbations}
\end{figure*}

The clean--noisy consistency experiment applies Gaussian noise with $\sigma=0.05$ to the current policy observation only. The goal observation, simulator dynamics, physical consequence, and success criterion remain clean. Each state is evaluated with three noise repeats. The archived smoke audit contains paired clean/noisy policy input images, confirming that the perturbation is applied at the policy input. Noise is added to the policy input image in uint8 range $[0,255]$ before ImageNet normalization, with independent per-pixel IID sampling and a fixed corruption seed of 123; the goal observation is kept clean. A strict corruption audit over all conditions reported zero corruption failures. The additional evaluation settings use pure additive brightness shifts of $+0.10$ and $+0.20$, and luminance-preserving saturation shifts of $+0.15$ and $+0.35$, all applied to the current policy input while leaving the goal and physical rollout clean. Figure~\ref{fig:pusht_visual_perturbations} visualizes these input-level transformations on one archived real PushT frame; it is illustrative and does not encode closed-loop success.

\subsection{Candidate-transition subspace erasure}
\begin{figure*}[!t]
    \centering
    \includegraphics[width=0.78\linewidth]{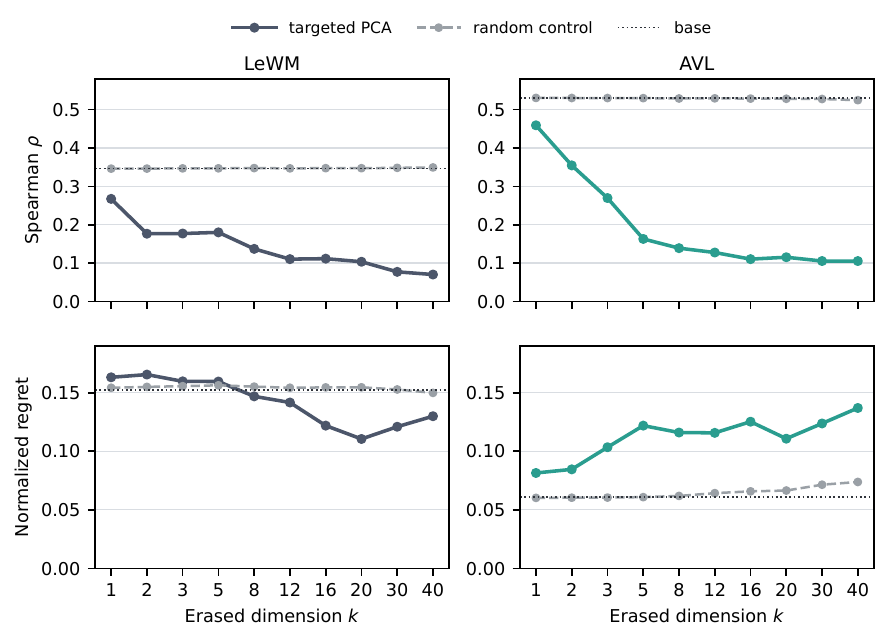}
    \caption{Candidate-transition subspace erasure sweep on TwoRoom for the directly computed dimensions $k\in\{1,2,3,5,8,12,16,20,30,40\}$. Targeted PCA removal is compared with same-dimensional random-subspace removal and the unmodified base.}
    \label{fig:targeted_erasure_sweep}
\end{figure*}

For each model, an operational PCA basis is fitted on 100 TwoRoom states to within-state centred predicted candidate-transition responses (the per-state mean over the candidates is subtracted, removing the component shared by all candidates of a state), with 5 action blocks per candidate. Before candidate scoring, the leading $k$-dimensional subspace is removed and the candidates are rescored by the norm of the erased predicted transition. Figure~\ref{fig:targeted_erasure_sweep} reports the extended, directly computed sweep $k\in\{1,2,3,5,8,12,16,20,30,40\}$ for LeWM and AVL, with 100 same-dimensional random-subspace repeats as control. Both branches are scored from the same candidate responses, with no interpolation or reconstructed aggregate values.

\subsection{Main experiment training conditions}
\label{app:main_training_conditions}
The main LeWM training run was performed on AutoDL with an NVIDIA RTX 5090 (32 GB), a 14-vCPU Intel Xeon Gold 6348 CPU at 2.60 GHz, and the project's `.venv` environment. The implementation used PyTorch 2.12.1+cu130 with its bundled CUDA 13.0 runtime, Lightning/stable-pretraining, bf16 mixed precision, Weights \& Biases, and a local CSV logger. The data were split into 90\% training and 10\% held-out samples; image resolution was $224$, embedding dimension was 192, history size was 3, and one future prediction was trained.

The optimizer was AdamW with learning rate $5\times10^{-5}$ and weight decay $10^{-3}$. Training used 10 epochs, batch size 128, 8 loader workers as specified by the training YAML, prefetch factor 3, pinned memory, persistent workers, bf16 precision, and gradient clipping at 1.0. The run summary recorded on the host additionally lists 12 workers; this is retained as an environment note rather than silently overwriting the versioned YAML setting. The SIGReg loss weight was 0.09, with 17 knots and 1024 projections. The checkpoint selected for evaluation was the final exported `lewm` checkpoint from the run; exact run identifiers and file hashes are retained in the accompanying experiment archive. These settings describe training-time configuration only; evaluation-time visual perturbations are applied separately at the policy input as specified in the perturbation protocol. All models use batch size 128. LeWM trains for 10 epochs, the action-grounded pathway for 10 epochs from scratch, and the vision-invariance pathway for 2 more epochs with the action head frozen. The standalone Vision-invariance baseline starts from the LeWM weights and trains for 2 epochs with the same clean-target objective but no action loss. Each model was trained with two random seeds.

\section{Additional Results}

\subsection{Complete experiment results}

\label{app:visual_conditions}

The remaining experimental results (the candidate-transition subspace erasure sweep and the additional visual-perturbation conditions) are reported below.

\subsection{Additional closed-loop rollout visualizations}

\begin{figure*}[!t]
\centering
\includegraphics[width=\linewidth]{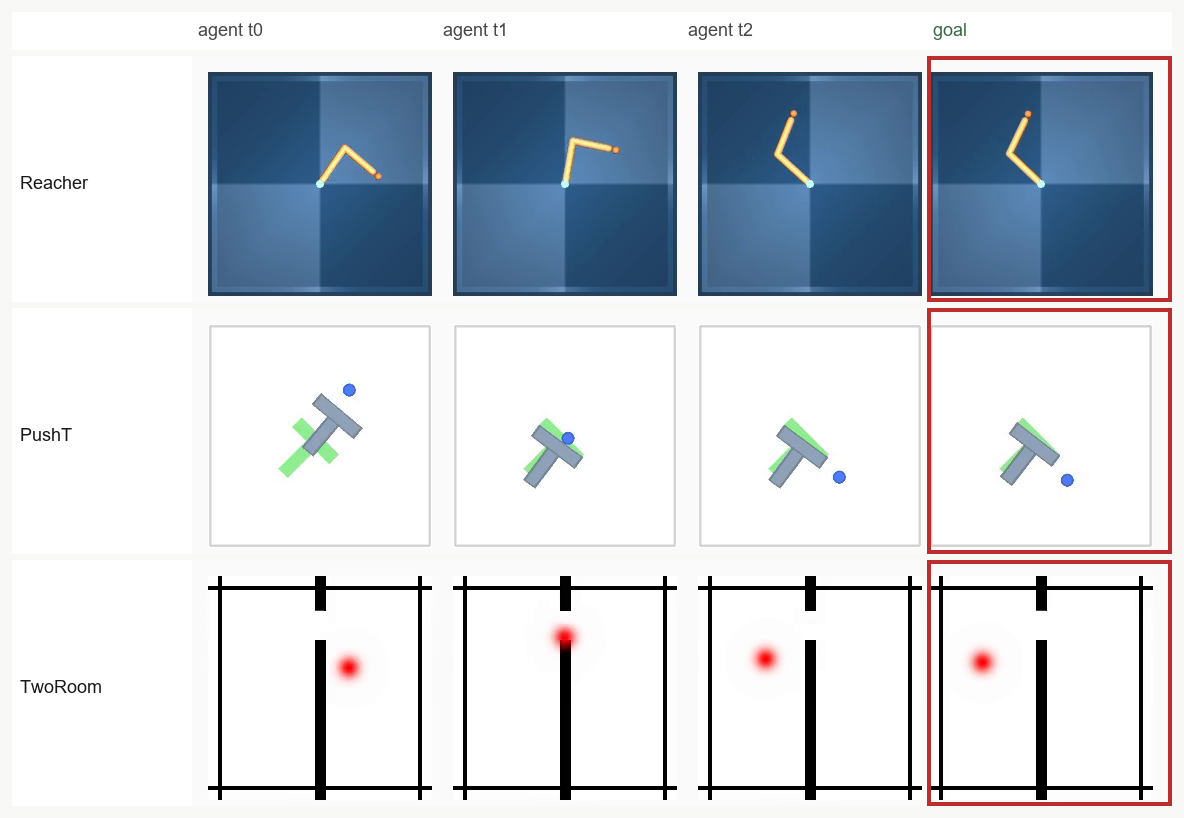}
\caption{Additional representative closed-loop rollouts for Reacher, PushT, and TwoRoom. Each row shows three temporally ordered agent observations and the corresponding goal observation.}
\label{fig:task_visualization_appendix}
\end{figure*}

Figure~\ref{fig:task_visualization_appendix} shows representative closed-loop rollouts from Reacher, PushT, and TwoRoom. These examples are provided as qualitative context; they are not matched head-to-head comparisons and do not constitute aggregate performance evidence.

\subsection{Analysis of the alignment and erasure results}
\label{app:mechanism_analysis}

\paragraph{Counterfactual alignment.}
On TwoRoom, predicted-to-physical Spearman correlation rises from $0.346$ (95\% bootstrap interval $[0.293,0.399]$) for \lewm{} to $0.508$ $[0.463,0.552]$ after action grounding and to $0.531$ $[0.486,0.574]$ after clean--perturbed alignment, while normalized regret falls from $0.152$ to $0.071$ and then to $0.061$: most of the gain comes from the action anchor, with a smaller but consistent increment from the vision-invariance stage. Exact Top-1 selection saturates at $47\%$ for both the action-grounded and the full AVL model, whereas Top-3 inclusion rises from $48\%$ to $77\%$, so the improvement lies mainly in bringing the physical oracle into a shortlist of candidates.

\paragraph{Erasure sweep.}
For AVL the signed change of Equation~\ref{eq:erasure_metrics} grows with the erased dimension ($\Delta R=+0.021$, $+0.024$, $+0.043$, $+0.061$, $+0.055$ for $k=1,2,3,5,8$) and its 95\% interval excludes zero at every $k$ we tested, while the same-dimensional random control stays below $0.015$: the leading transition directions are specifically, not merely dimensionally, necessary. For \lewm{} the same intervention does not hurt ($\Delta R\approx+0.01$ up to $k=5$) and becomes negative beyond ($-0.030$ at $k=16$ and $-0.042$ at $k=20$), so erasing its leading directions improves its regret. This asymmetry is the signature of the collapse: \lewm{}'s dominant directions are nuisance-dominated, whereas AVL's are the ones its consequence ranking relies on.

\section{Hyperparameters and Compute}

The main training configuration is summarized above. For reproducibility, the model uses an image size of 224, a 192 dimensional latent embedding, history size 3, and one prediction target. The run uses a 0.9 training split, AdamW with learning rate $5\times10^{-5}$ and weight decay $10^{-3}$, bf16 precision, gradient clipping at 1.0, batch size 128, prefetch factor 3, pinned memory, and persistent workers. The configuration-level worker count is 8; a separate runtime summary reports 12 workers. The planning evaluation uses the task-specific checkpoint exports and the CEM settings given in Section~\ref{sec:implementation_details}, with no training performed during evaluation.

\setlength{\abovetopsep}{2pt}
\setlength{\abovecaptionskip}{6pt}
\setlength{\aboverulesep}{0.25ex}
\setlength{\belowrulesep}{0.25ex}
\setlength{\floatsep}{3pt plus 1pt minus 1pt}
\setlength{\textfloatsep}{7pt plus 2pt minus 2pt}
\renewcommand{\arraystretch}{0.96}

\begin{table*}[!t]
\centering
\caption{Gaussian policy input noise, $\sigma=0.10$.}
\label{tab:gaussian010_results}
\begin{tabular*}{\linewidth}{@{\extracolsep{\fill}}lcccc@{}}
\toprule
Model & TwoRoom & PushT & Reacher & Cube \\
\midrule
\lewm{} & 65.7 $\pm$ 3.9 & 15.3 $\pm$ 5.8 & 42.6 $\pm$ 3.4 & 52.6 $\pm$ 7.2 \\
Action-grounded & 76.0 $\pm$ 1.6 & 10.0 $\pm$ 2.8 & 82.6 $\pm$ 3.3 & 64.2 $\pm$ 6.5 \\
Vision-invariance & 82.6 $\pm$ 2.9 & 31.3 $\pm$ 4.1 & \textbf{84.0 $\pm$ 2.8} & 68.0 $\pm$ 9.2 \\
AVL (ours) & \textbf{88.7 $\pm$ 1.9} & \textbf{81.5 $\pm$ 4.1} & 78.2 $\pm$ 4.4 & \textbf{72.7 $\pm$ 7.4} \\
\bottomrule
\end{tabular*}
\end{table*}

\begin{table*}[!t]
\centering
\caption{Brightness shift, $+0.20$.}
\label{tab:brightness020_results}
\begin{tabular*}{\linewidth}{@{\extracolsep{\fill}}lcccc@{}}
\toprule
Model & TwoRoom & PushT & Reacher & Cube \\
\midrule
\lewm{} & 78.0 $\pm$ 4.8 & 39.6 $\pm$ 3.3 & 11.6 $\pm$ 6.9 & 45.4 $\pm$ 3.7 \\
Action-grounded & 86.0 $\pm$ 1.4 & 30.5 $\pm$ 2.6 & 7.3 $\pm$ 0.9 & 42.0 $\pm$ 1.6 \\
Vision-invariance & 80.7 $\pm$ 3.7 & 62.6 $\pm$ 2.4 & 14.1 $\pm$ 6.6 & 48.5 $\pm$ 4.9 \\
AVL (ours) & \textbf{88.9 $\pm$ 2.4} & \textbf{72.0 $\pm$ 0.7} & \textbf{14.7 $\pm$ 8.1} & \textbf{48.7 $\pm$ 5.7} \\
\bottomrule
\end{tabular*}
\end{table*}

\begin{table*}[!t]
\centering
\caption{Saturation shift, $+0.35$.}
\label{tab:saturation035_results}
\begin{tabular*}{\linewidth}{@{\extracolsep{\fill}}lcccc@{}}
\toprule
Model & TwoRoom & PushT & Reacher & Cube \\
\midrule
\lewm{} & 54.8 $\pm$ 4.5 & 14.0 $\pm$ 3.9 & 24.4 $\pm$ 2.4 & 70.8 $\pm$ 5.2 \\
Action-grounded & 36.0 $\pm$ 1.1 & 31.3 $\pm$ 1.8 & 54.0 $\pm$ 4.3 & 69.3 $\pm$ 7.4 \\
Vision-invariance & 85.3 $\pm$ 3.4 & 22.4 $\pm$ 1.6 & 49.5 $\pm$ 2.4 & 71.4 $\pm$ 3.8 \\
AVL (ours) & \textbf{86.7 $\pm$ 3.4} & \textbf{62.0 $\pm$ 4.2} & \textbf{65.3 $\pm$ 0.9} & \textbf{74.0 $\pm$ 1.6} \\
\bottomrule
\end{tabular*}
\end{table*}

\begin{table*}[!t]
\centering
\caption{Gaussian policy input noise, $\sigma=0.03$.}
\label{tab:app_g003}
\begin{tabular*}{\linewidth}{@{\extracolsep{\fill}}lcccc@{}}
\toprule
Model & TwoRoom & PushT & Reacher & Cube \\
\midrule
\lewm{} & 83.5 $\pm$ 2.6 & 85.6 $\pm$ 1.7 & 80.2 $\pm$ 4.3 & 72.5 $\pm$ 8.2 \\
Action-grounded & 90.7 $\pm$ 2.9 & 92.0 $\pm$ 1.6 & \textbf{83.1 $\pm$ 5.2} & 71.3 $\pm$ 8.9 \\
Vision-invariance & 81.4 $\pm$ 3.4 & 92.6 $\pm$ 3.4 & 82.8 $\pm$ 4.1 & 70.9 $\pm$ 9.8 \\
AVL (ours) & \textbf{92.0 $\pm$ 1.6} & \textbf{96.7 $\pm$ 0.9} & 80.5 $\pm$ 5.8 & \textbf{74.6 $\pm$ 7.3} \\
\bottomrule
\end{tabular*}
\end{table*}

\begin{table*}[!t]
\centering
\caption{Gaussian policy input noise, $\sigma=0.05$.}
\label{tab:app_g005}
\begin{tabular*}{\linewidth}{@{\extracolsep{\fill}}lcccc@{}}
\toprule
Model & TwoRoom & PushT & Reacher & Cube \\
\midrule
\lewm{} & 83.3 $\pm$ 1.8 & 65.3 $\pm$ 4.8 & 73.9 $\pm$ 2.7 & 71.3 $\pm$ 9.2 \\
Action-grounded & 90.8 $\pm$ 1.6 & 74.7 $\pm$ 5.5 & 82.0 $\pm$ 2.4 & 68.7 $\pm$ 8.9 \\
Vision-invariance & 83.1 $\pm$ 1.8 & 91.2 $\pm$ 0.9 & \textbf{87.4 $\pm$ 2.0} & 69.3 $\pm$ 8.2 \\
AVL (ours) & \textbf{92.0 $\pm$ 1.6} & \textbf{95.3 $\pm$ 0.9} & 82.5 $\pm$ 3.9 & \textbf{74.4 $\pm$ 9.4} \\
\bottomrule
\end{tabular*}
\end{table*}

\begin{table*}[!t]
\centering
\caption{Brightness shift, $-0.10$.}
\label{tab:app_brightn010}
\begin{tabular*}{\linewidth}{@{\extracolsep{\fill}}lcccc@{}}
\toprule
Model & TwoRoom & PushT & Reacher & Cube \\
\midrule
\lewm{} & 83.6 $\pm$ 4.6 & 87.6 $\pm$ 3.8 & 17.6 $\pm$ 3.5 & 44.0 $\pm$ 7.4 \\
Action-grounded & 92.0 $\pm$ 4.3 & 93.0 $\pm$ 2.9 & \textbf{31.3 $\pm$ 0.9} & 44.7 $\pm$ 6.8 \\
Vision-invariance & 81.3 $\pm$ 2.9 & 92.0 $\pm$ 2.8 & 21.4 $\pm$ 4.2 & 44.1 $\pm$ 5.5 \\
AVL (ours) & \textbf{92.7 $\pm$ 2.4} & \textbf{96.6 $\pm$ 1.7} & 30.0 $\pm$ 5.7 & \textbf{46.0 $\pm$ 7.1} \\
\bottomrule
\end{tabular*}
\end{table*}

\begin{table*}[!t]
\centering
\caption{Brightness shift, $+0.10$.}
\label{tab:app_brightp010}
\begin{tabular*}{\linewidth}{@{\extracolsep{\fill}}lcccc@{}}
\toprule
Model & TwoRoom & PushT & Reacher & Cube \\
\midrule
\lewm{} & 82.8 $\pm$ 1.3 & 91.2 $\pm$ 3.8 & 15.6 $\pm$ 4.7 & 56.8 $\pm$ 4.3 \\
Action-grounded & 90.6 $\pm$ 0.9 & 92.0 $\pm$ 3.4 & 20.0 $\pm$ 4.3 & \textbf{63.6 $\pm$ 6.8} \\
Vision-invariance & 81.3 $\pm$ 4.7 & 91.3 $\pm$ 2.4 & 18.1 $\pm$ 2.8 & 55.2 $\pm$ 5.2 \\
AVL (ours) & \textbf{92.7 $\pm$ 0.9} & \textbf{92.0 $\pm$ 2.2} & \textbf{24.7 $\pm$ 1.9} & 54.1 $\pm$ 1.6 \\
\bottomrule
\end{tabular*}
\end{table*}

\begin{table*}[!t]
\centering
\caption{Brightness shift, $-0.20$.}
\label{tab:app_brightn020}
\begin{tabular*}{\linewidth}{@{\extracolsep{\fill}}lcccc@{}}
\toprule
Model & TwoRoom & PushT & Reacher & Cube \\
\midrule
\lewm{} & 54.0 $\pm$ 4.8 & 12.0 $\pm$ 3.6 & 14.8 $\pm$ 8.2 & 37.6 $\pm$ 5.6 \\
Action-grounded & 43.2 $\pm$ 4.9 & 30.5 $\pm$ 3.5 & 20.3 $\pm$ 9.0 & 44.7 $\pm$ 4.1 \\
Vision-invariance & 82.0 $\pm$ 3.2 & 22.6 $\pm$ 0.9 & 22.5 $\pm$ 6.2 & 42.9 $\pm$ 4.3 \\
AVL (ours) & \textbf{91.3 $\pm$ 3.4} & \textbf{40.7 $\pm$ 1.3} & \textbf{24.7 $\pm$ 5.3} & \textbf{47.3 $\pm$ 4.7} \\
\bottomrule
\end{tabular*}
\end{table*}

\begin{table*}[!t]
\centering
\caption{Saturation shift, $+0.15$.}
\label{tab:app_satp015}
\begin{tabular*}{\linewidth}{@{\extracolsep{\fill}}lcccc@{}}
\toprule
Model & TwoRoom & PushT & Reacher & Cube \\
\midrule
\lewm{} & 83.2 $\pm$ 1.8 & 91.6 $\pm$ 3.2 & 56.4 $\pm$ 6.5 & 70.2 $\pm$ 6.1 \\
Action-grounded & 92.0 $\pm$ 1.6 & 78.0 $\pm$ 1.8 & \textbf{85.3 $\pm$ 5.3} & 72.0 $\pm$ 5.9 \\
Vision-invariance & 83.1 $\pm$ 1.9 & 85.3 $\pm$ 5.2 & 81.6 $\pm$ 5.7 & 69.6 $\pm$ 5.4 \\
AVL (ours) & \textbf{94.0 $\pm$ 1.6} & \textbf{93.6 $\pm$ 3.7} & 82.7 $\pm$ 3.8 & \textbf{73.7 $\pm$ 5.7} \\
\bottomrule
\end{tabular*}
\end{table*}

\clearpage
\end{document}